\documentclass[runningheads]{llncs}
\usepackage{eccv}

\usepackage{eccvabbrv}
\usepackage{graphicx}
\usepackage{booktabs}
\usepackage{multirow}
\usepackage{xcolor}
\usepackage{colortbl}
\usepackage[accsupp]{axessibility}
\usepackage[pagebackref,breaklinks,colorlinks,citecolor=eccvblue]{hyperref}
\usepackage{orcidlink}

\begin{document}

\title{AerialVLA: A Vision-Language-Action Model for UAV Navigation via Minimalist End-to-End Control} 
\titlerunning{AerialVLA}

\author{Peng Xu\inst{1}\orcidlink{0009-0002-2523-0614} \and
Zhengnan Deng\inst{1}\orcidlink{0009-0003-8124-9202} \and
Jiayan Deng\inst{1}\orcidlink{0009-0007-9030-3206} \and
Zonghua Gu\inst{2}\orcidlink{0000-0003-4228-2774} \and
Shaohua Wan\inst{1}\orcidlink{0000-0001-7013-9081}\thanks{Corresponding author.}}

\authorrunning{P. Xu et al.}

\institute{Shenzhen Institute for Advanced Study, University of Electronic Science and Technology of China, Shenzhen, China\\
\email{pxu023@gmail.com, shaohua.wan@uestc.edu.cn} \and
Department of Computer Science, Hofstra University, Hempstead, USA\\
\email{zonghua.gu@hofstra.edu}}

\maketitle

\begin{abstract}
  Vision-Language Navigation (VLN) for Unmanned Aerial Vehicles (UAVs) demands complex visual interpretation and continuous control in dynamic 3D environments. Existing hierarchical approaches rely on dense oracle guidance or auxiliary object detectors, creating semantic gaps and limiting genuine autonomy. We propose AerialVLA, a minimalist end-to-end Vision-Language-Action framework mapping raw visual observations and fuzzy linguistic instructions directly to continuous physical control signals. First, we introduce a streamlined dual-view perception strategy that reduces visual redundancy while preserving essential cues for forward navigation and precise grounding, which additionally facilitates future simulation-to-reality transfer. To reclaim genuine autonomy, we deploy a fuzzy directional prompting mechanism derived solely from onboard sensors, completely eliminating the dependency on dense oracle guidance. Ultimately, we formulate a unified control space that integrates continuous 3-Degree-of-Freedom (3-DoF) kinematic commands with an intrinsic landing signal, freeing the agent from external object detectors for precision landing. Extensive experiments on the TravelUAV benchmark demonstrate that AerialVLA achieves state-of-the-art performance in seen environments. Furthermore, it exhibits superior generalization in unseen scenarios by achieving nearly three times the success rate of leading baselines, validating that a minimalist, autonomy-centric paradigm captures more robust visual-motor representations than complex modular systems. Code is available at: \url{https://github.com/XuPeng23/AerialVLA}

  \keywords{Vision-Language Navigation \and Vision-Language-Action \and UAV Control \and End-to-End Learning}
\end{abstract}

\begin{figure}[t]
  \centering
  \includegraphics[width=\linewidth]{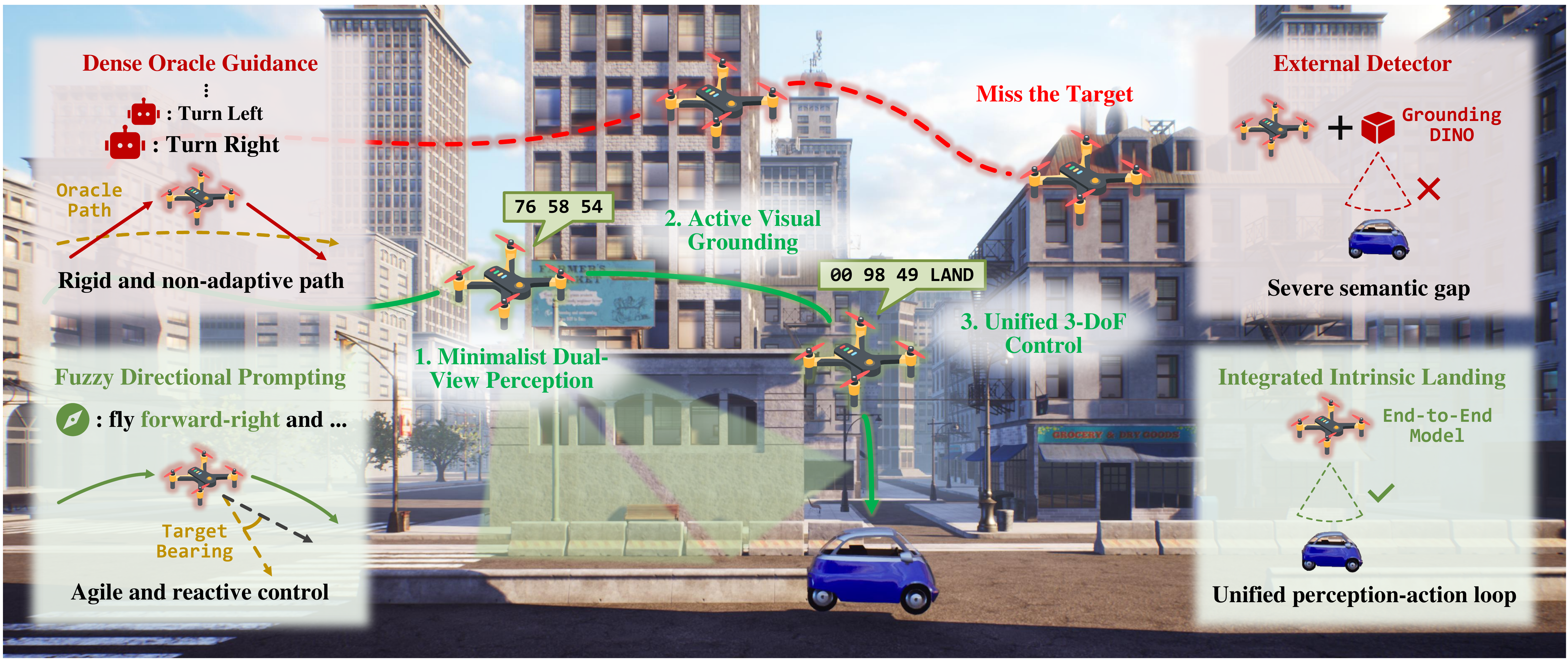}
  \caption{\textbf{Comparison of UAV navigation paradigms.} While existing modular methods (red) rely on oracle guidance and external detectors, AerialVLA (green) achieves agile autonomous navigation and precise landing via a unified end-to-end policy driven by fuzzy onboard hints and intrinsic stopping.}
  \label{fig:figure_teaser}
\end{figure}

\section{Introduction}
\label{sec:intro}
Vision-Language Navigation (VLN) has seen remarkable progress in ground-based agents. However, extending VLN to Unmanned Aerial Vehicles (UAVs) introduces unique challenges due to the complexity of 3D open-world environments. This transition marks a paradigm shift in aerial robotics, as underscored by recent surveys~\cite{sapkota2025uavs,tian2025uavs,javaid2024large}, evolving from automated flight to agentic UAVs capable of autonomous perception, reasoning, and interaction. The field has rapidly advanced alongside the development of specialized benchmarks, ranging from object goal navigation~\cite{xiao2025uav} to high-level spatial reasoning~\cite{zhang2025spatialsky, ferrag2025uavbench} and realistic flight simulation~\cite{gao2026openfly, wang2025towards}. While this flourishing ecosystem has established rigorous standards for evaluating the cognitive understanding of a scene, the challenge of mastering continuous 6-Degree-of-Freedom (6-DoF) physical execution remains arduous.

Unlike ground robots constrained to a 2D plane, UAVs must navigate in a full 6-DoF state space, managing 3D position and orientation, while interpreting visual cues from actively changing ego-centric viewpoints. This introduces unique challenges in continuous control under gravitational and inertial constraints. The goal of UAV-VLN is to enable drones to navigate to a natural-language-described target autonomously. This capability is critical for applications ranging from search and rescue to remote inspection, where GPS signals may be unreliable or target coordinates are unknown.

Despite this potential, existing UAV-VLN approaches often suffer from a reliance on what we term ``double crutches'', which limits their autonomy in the wild. A primary issue is the dependency on oracle guidance. State-of-the-art benchmarks such as TravelUAV~\cite{wang2025towards} and recent methods~\cite{jiang2025longfly, zhang2026embodied} typically inject dense, ground-truth-derived directional hints (e.g., ``Turn Right'') directly into the input prompts. This practice effectively degrades the navigation agent into a passive instruction follower, bypassing the core challenge of active spatial reasoning and path planning. Furthermore, these modular pipelines frequently necessitate an external object detector, such as Grounding DINO~\cite{liu2024grounding}, to trigger the landing phase due to a lack of fine-grained visual grounding. This dependency creates a disjointed perception-control loop where the policy learns how to move but relies on an external black box to decide when to stop, thereby reducing system robustness when detectors fail in open-world scenarios.

As illustrated in Figure~\ref{fig:figure_teaser}, to address the aforementioned double crutches, we present \textbf{AerialVLA}, a minimalist end-to-end Vision-Language-Action framework. Unlike existing modular approaches (red trajectory) that suffer from rigid paths and semantic gaps due to dense oracle guidance and external detectors, AerialVLA (green trajectory) establishes a unified perception-action loop. By replacing exact oracles with fuzzy onboard directional hints, our agent is driven to perform active visual grounding, enabling agile and robust spatial reasoning. Concurrently, by mapping raw observations directly to continuous physical signals, AerialVLA unlocks an integrated intrinsic landing, seamlessly unifying long-distance cruising and precision stopping within a single policy.

In summary, our main contributions are three-fold:
\begin{itemize}
    \item \textbf{Minimalist Dual-View Perception:} We fuse front and down views into a streamlined visual interface aligned with consumer UAV hardware. This design discards multi-camera redundancies while preserving the essential geometric and semantic cues required for forward navigation and precise target grounding.
    \item \textbf{Fuzzy Directional Prompting:} We eliminate the reliance on step-by-step oracle guidance by introducing fuzzy directional prompts. Derived solely from onboard IMU estimations, this formulation accommodates real-world localization uncertainty and forces the agent to learn robust, active spatial reasoning rather than passive instruction following.
    \item \textbf{High-DoF Control via Numerical Tokenization:} We tokenize a continuous 3-DoF action space compatible with standard UAV control APIs, effectively leveraging the pre-trained numerical reasoning of LLMs. This unlocks an end-to-end intrinsic stopping policy, unifying cruising and precision landing without external object detectors.
\end{itemize}

\section{Related Work}
\subsection{Benchmarks and Datasets for Aerial Agents}
High-quality simulators drive aerial autonomy. AerialVLN~\cite{liu2023Aerialvln} and AVDN~\cite{fan2023aerial} pioneered instruction-following and dialog tasks, while UAV-ON~\cite{xiao2025uav} introduced open-world object navigation. Recent focus expanded to cognitive capabilities: SpatialSky-Bench~\cite{zhang2025spatialsky} evaluates spatial reasoning, and UAVBench~\cite{ferrag2025uavbench} assesses mission planning. However, these benchmarks primarily evaluate high-level perception or abstract logic. While OpenFly~\cite{gao2026openfly} offers diverse trajectories via discrete macro-actions, TravelUAV~\cite{wang2025towards} focuses on 3D continuous navigation precision. We adopt TravelUAV to rigorously evaluate our agent's continuous maneuvering capabilities.

\subsection{Vision-Language Navigation for UAVs}
As a critical domain of embodied intelligence, the UAV-VLN task requires the integration of multimodal perception, spatial reasoning, and motion planning in complex 3D environments~\cite{yao2025aeroverse}. Unlike prior surveys that categorize methods by algorithmic foundations, we classify existing approaches based on their control granularity to explicitly highlight end-to-end execution challenges.

\textbf{High-Level Planning and Waypoint Prediction.} Most works, including early baselines like CMA~\cite{Anderson_2018_CVPR} and AVDN~\cite{fan2023aerial}, formulate navigation as a planning problem, predicting spatial waypoints for low-level controllers (e.g., PID). These includes modular systems (CityNavAgent~\cite{zhang2025citynavagent}, SkyVLN~\cite{li2025skyvln}, AeroDuo~\cite{wu2025aeroduo}) generating long-horizon or collaborative plans, and mission-generation frameworks (UAV-VLA~\cite{sautenkov2025uav}). Learning-based policies (TravelUAV~\cite{wang2025towards}, OpenVLN~\cite{lin2025openvln}, LongFly~\cite{jiang2025longfly}, NavFoM~\cite{zhang2026embodied}, FlightGPT~\cite{cai2025flightgpt}) regress waypoints, targets, or trajectories via fine-tuned foundation models and spatiotemporal contexts, while ANWM~\cite{zhang2025aerial} evaluates candidate trajectories using a generative world model. While effective for task decomposition, decoupling perception from control discards fine-grained visual cues critical for aerodynamic stability, introduces severe inference latency, and forces reliance on semantically unaware low-level controllers.

\textbf{Training-Free and Reasoning-Centric Approaches.} A parallel trend exploits frozen foundation models~\cite{team2023gemini, openai2023gpt4} to bypass policy training. For instance, SPF~\cite{hu2025see} prompts them to predict 2D waypoints for heuristic 3D unprojection, which often violates physical constraints due to absent depth perception. Meanwhile, TypeFly~\cite{chen2025typefly} generates code primitives for task planning. Despite their strong zero-shot capabilities, the sequential token generation of these massive LLMs creates inference latency fundamentally incompatible with real-time aerial agility.

\textbf{End-to-End Continuous Control.} This paradigm maps visual observations directly to continuous control signals (e.g., velocity, thrust). RaceVLA~\cite{serpiva2025racevla} and CognitiveDrone~\cite{lykov2025cognitivedrone} pioneer this direction by adapting fine-tuned VLA models for flight. Nevertheless, these works are primarily confined to structured environments like racing tracks, simplifying navigation into discrete gate-selection tasks. While CognitiveDrone adds an auxiliary VLM, this decoupled, lower-frequency reasoning remains separated from the high-frequency control loop. Lacking intrinsic and unified spatial reasoning, these models struggle with the semantic grounding required for unstructured open-world exploration.

To overcome the dichotomy between decoupled hierarchical planning and structurally confined continuous control, AerialVLA introduces a minimalist VLA paradigm. By unifying fuzzy directional prompting with an intrinsic landing policy, we achieve robust, open-world reasoning and precise maneuvering without relying on external detectors or heavy reasoning chains.

\begin{figure}[t]
  \centering
  \includegraphics[width=\linewidth]{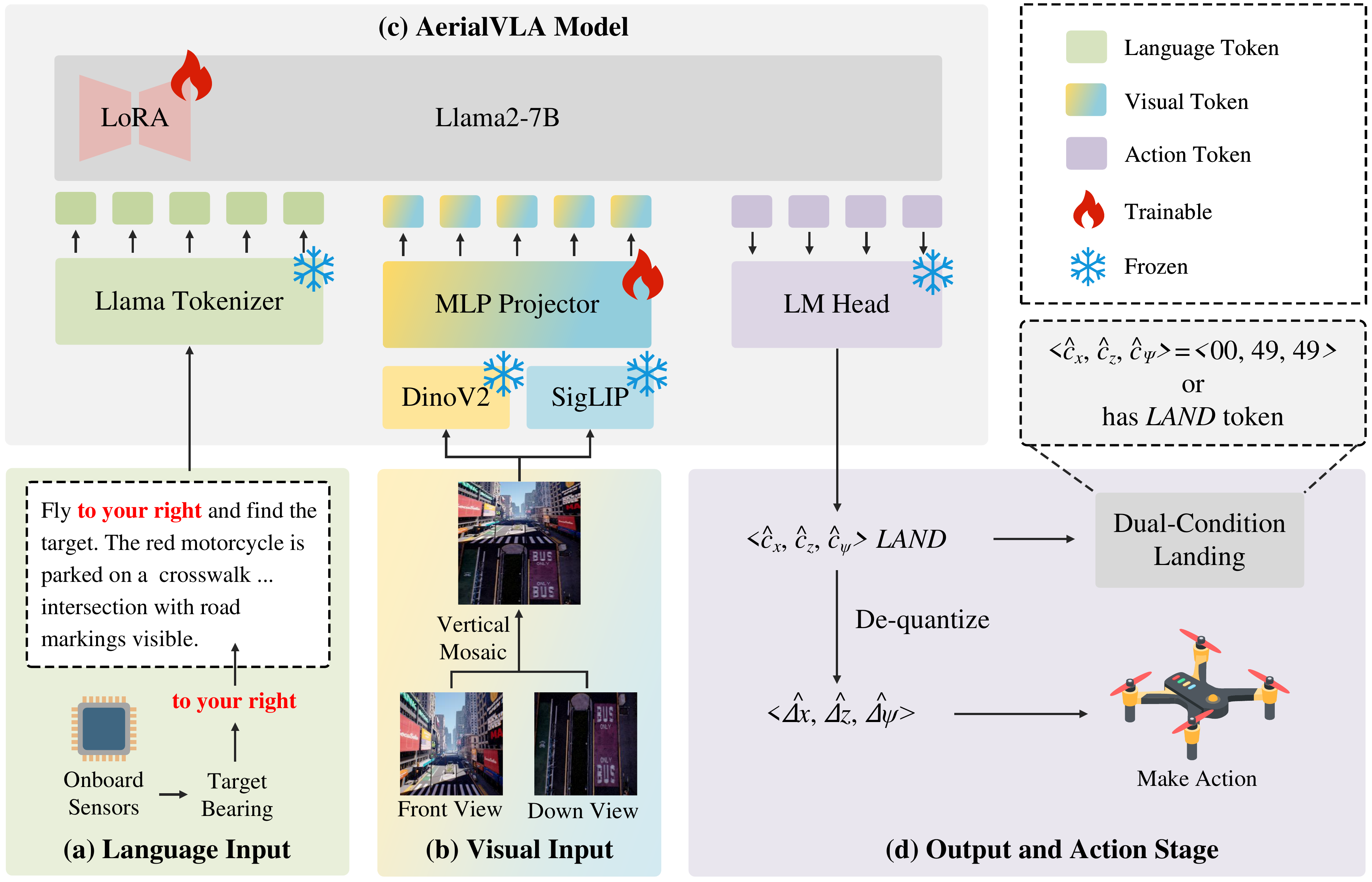}
  \caption{\textbf{The architecture of AerialVLA.} The framework processes multimodal inputs to generate continuous control signals end-to-end. \textbf{(a) Language Input} constructs prompts with fuzzy directional hints derived from the IMU, eliminating oracle reliance. \textbf{(b) Visual Input} fuses front and down views via a vertical mosaic. \textbf{(c) The AerialVLA Model} utilizes a Llama-2 backbone with LoRA to autoregressively predict numerical tokens. \textbf{(d) Output and Action Stage} decodes tokens into spatial offsets for velocity control, or triggers the dual-condition landing.}
  \label{fig:figure_pipeline}
\end{figure}

\subsection{Vision-Language-Action Models: From Ground to Air}
The paradigm of Vision-Language-Action (VLA) models has revolutionized robotic control by unifying perception and action within a single transformer. Generalist models like RT-2~\cite{zitkovich2023rt} and OpenVLA~\cite{kim2025openvla} demonstrate remarkable generalization by fine-tuning VLMs to output robot actions directly as language tokens. While end-to-end learning has been successfully applied to ground navigation~\cite{shah2023vint, tian2024drivevlm, ku2020room} to bypass traditional explicit map construction~\cite{cadena2017past}, existing VLA successes remain predominantly focused on quasi-static manipulation tasks or 2D ground rovers.

Extending the VLA paradigm to aerial platforms introduces fundamental challenges absent in ground-based settings. First, UAVs operate in a fully dynamic 6-DoF state space subject to continuous gravitational and inertial forces, where pitch and roll dynamics are tightly coupled with translational motion. Second, unlike reversible manipulation actions, flight control requires managing tight kinematic constraints where failure is often terminal (e.g., collision). AerialVLA bridges this gap by adapting the VLA architecture specifically for aerial embodiment. We demonstrate that a direct token-prediction approach, when grounded in large-scale expert demonstrations, effectively masters the continuous 3-DoF control required for high-stakes, open-world aerial navigation.

\section{Method}
\label{sec:method}
\subsection{Overview}
We present \textbf{AerialVLA}, an end-to-end Vision-Language-Action framework designed to endow Unmanned Aerial Vehicles (UAVs) with autonomous navigation capabilities in open-world environments. As illustrated in Figure~\ref{fig:figure_pipeline}, our architecture is built upon the OpenVLA-7B~\cite{kim2025openvla} backbone, which unifies a hybrid visual encoder with a Llama 2~\cite{touvron2023llama} language model using the Transformer architecture~\cite{vaswani2017attention}. To adapt this generalist model for aerial navigation in dynamic 6-DoF state space, we introduce three key innovations: (1) a streamlined \textbf{Dual-View Perception} interface that balances informational gain with computational efficiency; (2) a \textbf{Fuzzy Directional Prompting} mechanism eliminating reliance on oracle assistance; and (3) a \textbf{Numerical Action Tokenization} strategy that leverages the pre-trained numerical reasoning of LLMs for precise control.

\subsection{Minimalist Dual-View Perception}
\label{sec:perception}
While recent UAV-VLN benchmarks typically simulate multi-camera arrays (e.g., five distinct views), processing such high-dimensional inputs significantly increases computational overhead and inference latency, hindering real-time deployment. Moreover, redundant views often contribute marginal utility for forward-facing tasks. To address this, we propose a minimalist dual-view strategy that aligns with consumer UAV hardware configurations. We explicitly process only the front and down views, where the front view provides critical cues for obstacle avoidance and target identification, while the down view is essential for precise ground alignment and landing maneuvers.

Formally, given the \textit{front} image $I_{\text{front}} \in \mathbb{R}^{H \times W \times 3}$ and \textit{down} image $I_{\text{down}} \in \mathbb{R}^{H \times W \times 3}$ from the AirSim simulator~\cite{Shah2018AirSim}, we perform vertical concatenation to form a composite observation $I_{\text{comp}} = [I_{\text{front}}; I_{\text{down}}] \in \mathbb{R}^{2H \times W \times 3}$. This composite is resized to the required $224 \times 224$ resolution and processed by a hybrid visual encoder combining SigLIP~\cite{Zhai_2023_ICCV} and DINOv2~\cite{oquab2024dinov}. Both encoders utilize the Vision Transformer (ViT)~\cite{dosovitskiy2021an} architecture, with SigLIP providing language-aligned semantic features via CLIP-style~\cite{radford2021learning} objectives and DINOv2 capturing robust, fine-grained spatial representations. To seamlessly map this dual-view input into the LLM embedding space, we fully fine-tune the visual projector. This adaptation ensures that the vertically compressed inputs effectively preserve crucial navigation cues, such as object categories and terrain textures. Furthermore, the default $90^\circ$ Field-of-View (FOV) in AirSim ensures the front and down views naturally align at their horizontal boundary, establishing physical and semantic continuity. Significantly, the $224 \times 224$ input dimension inherently aligns the stitching seam with the non-overlapping $14 \times 14$ patch grids of the ViT encoders. This architectural alignment prevents intra-patch corruption, enabling the self-attention mechanism to cleanly resolve cross-view spatial relationships.

\subsection{Fuzzy Directional Prompting}
\label{sec:prompt}
A major limitation of existing UAV-VLN methods (e.g., TravelUAV~\cite{wang2025towards}) is their reliance on dense, step-by-step oracle guidance derived from pre-recorded optimal trajectories. This compromises autonomy by degrading the agent into a passive instruction follower. AerialVLA removes this dependency by constructing instruction prompts using only a fuzzy directional hint derived from onboard sensors (IMU/GPS). Specifically, we define a mapping function $\mathcal{M}$ that discretizes the relative bearing $\theta$ of the target into coarse-grained semantic buckets, as detailed in Table~\ref{tab:fuzzy_hints}. By deliberately stripping away these dense oracles, we subject our agent to a fundamentally more challenging and rigorous learning objective. Rather than providing exact step-by-step trajectory alignments, our input offers only coarse directional priors, introducing significant spatial ambiguity relative to the ground-truth path. Importantly, this coarse-grained formulation not only aligns with real-world imperfect localization but also introduces necessary tolerance during training. By providing directional guidance without precise angular resolution, it forces the model to rely primarily on active visual grounding, thereby enhancing policy robustness against sensor noise and environmental ambiguity.

\begin{table}[tbp]
\centering
\caption{Fuzzy directional hint mapping from onboard IMU-derived relative angle.}
\label{tab:fuzzy_hints}
\small
\begin{tabular}{cc}
\toprule
Angle Range $|\theta|$ & Fuzzy Hint ($\theta > 0$ / $\theta < 0$) \\
\midrule
$0^\circ \le |\theta| \le 15^\circ$ & \multicolumn{1}{c}{``straight ahead''} \\
$15^\circ < |\theta| \le 60^\circ$ & ``forward-right'' / ``forward-left'' \\
$60^\circ < |\theta| \le 120^\circ$ & ``to your right'' / ``to your left'' \\
$120^\circ < |\theta| \le 180^\circ$ & ``to your right rear'' / ``to your left rear'' \\
\bottomrule
\end{tabular}
\end{table}

\begin{figure}[tbp]
  \centering
  \includegraphics[width=0.9\linewidth]{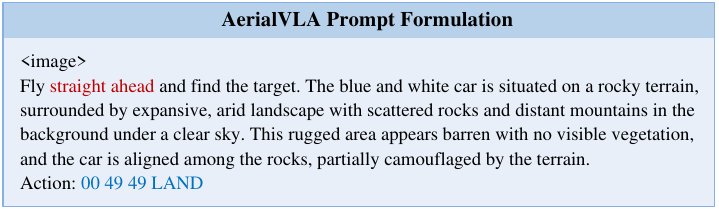}
  \caption{AerialVLA prompt formulation. The structured prompt comprises four components: (i) an \texttt{<image>} token for visual input, (ii) a fuzzy directional hint (red), (iii) a detailed target description, and (iv) the corresponding numerical control actions (blue).}
  \label{fig:figure_prompt}
\end{figure}

As illustrated in Figure~\ref{fig:figure_prompt}, the prompt is structured as a natural sequence commencing with the visual placeholder, followed by the directional hint and the detailed target description. This format integrates multimodal context into a cohesive narrative for the LLM. Distinctively, AerialVLA executes a fully reactive policy based strictly on the current observation and immediate hint. By intentionally bypassing complex spatiotemporal memory buffers, this minimalist design drastically reduces inference latency and prevents the accumulation of cascading errors from stale states. Ultimately, prioritizing this instantaneous agility establishes a highly robust foundation for reactive visual-motor control in dynamic environments.

\textbf{Geometry-Consistent Supervision.} Training an autonomous policy on fuzzy prompts requires strictly resolving mathematical ambiguities in expert demonstrations. Modular baselines utilize dense oracle guidance, providing sufficient local context to justify detour maneuvers. Conversely, AerialVLA relies solely on coarse hints. Pairing a lateral target hint with a straight-flight label in an obstacle-free environment introduces causal confusion, which typically stems from delayed human pilot reactions. To resolve this ambiguity while preserving critical collision avoidance skills, we propose a geometry-consistent filtering strategy using lateral depth maps $d_{\text{lat}}$. Formally, we evaluate frames exhibiting a significant lateral target bearing ($|\theta| > 60^\circ$) coupled with a near-zero expert yaw rate ($\omega_\psi \approx 0$). For these ambiguous frames, we inspect the corresponding side-view depth: if the lateral space is clear ($\min(d_{\text{lat}}) > 20$m), we discard the sample. However, if a nearby obstacle is detected ($\min(d_{\text{lat}}) \le 20$m), we retain the straight flight label as a valid, essential evasion maneuver. This geometric curation removes approximately 4\% of the total training frames, guaranteeing that AerialVLA learns to navigate structural constraints, such as urban intersections, without being penalized by contradictory supervision.

\subsection{High-DoF Control via Numerical Tokenization}
\label{sec:action}
\textbf{Task-Oriented Action Space and Unified Landing.} We define a continuous 3-DoF action space $\mathcal{A} = \langle \Delta x, \Delta z, \Delta \psi \rangle$ to decouple altitude, forward progression, and heading adjustments. This formulation delegates low-level stabilization (e.g., roll/pitch dynamics) to the flight controller, aligning with high-level motion primitives exposed by commercial UAV APIs (e.g., DJI SDK, PX4) to facilitate real-world deployment. Crucially, unlike modular baselines that rely on external object detectors (e.g., Grounding DINO~\cite{liu2024grounding}) to trigger flight termination, AerialVLA learns an intrinsic stopping policy. During training, we explicitly align terminal frames with a zero-displacement label vector $\langle 0, 0, 0 \rangle$ and the standard text token \texttt{LAND}. Consequently, landing is executed end-to-end via a robust dual-condition check: either the generation of the \texttt{LAND} token or the prediction of near-zero spatial offsets. This unifies navigation and landing into a single behavior cloning objective, enabling the agent to autonomously trigger flight termination upon visual convergence.

\textbf{Standard Numerical Tokenization.} 
Previous VLA approaches like RT-2~\cite{zitkovich2023rt} typically introduce special action tokens (e.g., \texttt{<act\_0>} to \texttt{<act\_255>}). In data-constrained UAV domains, training these new embeddings from scratch creates a severe cold-start problem, forcing the model to re-learn basic ordinal relationships. Instead, AerialVLA maps continuous action dimensions, discretized into $N=99$ bins, directly to existing numerical tokens within the LLM vocabulary. This elegantly leverages the understanding of magnitude and order inherent in the pre-trained model, yielding faster convergence and smoother control trajectories. At each timestep $t$, the model autoregressively predicts three integer tokens:
\begin{equation}
    \langle \hat{c}_x, \hat{c}_z, \hat{c}_\psi \rangle = \text{LLM}(E_{\text{vis}}, E_{\text{prompt}})
\end{equation}
where $\hat{c}_k \in \{0, 1, \dots, 98\}$. These categorical indices are deterministically de-quantized into continuous physical commands $\langle \hat{\Delta x}, \hat{\Delta z}, \hat{\Delta \psi} \rangle$. Specifically, $\hat{\Delta x} \in [0, 5]$ and $\hat{\Delta z} \in [-5, 5]$ represent the forward and vertical displacements in meters, while $\hat{\Delta \psi} \in [-\pi, \pi]$ denotes the yaw change in radians. These action boundaries are strictly derived from the statistical distribution of the expert dataset. To ensure robust physical execution, we map these spatial offsets to continuous velocity commands. Rather than relying on default position-control APIs that often induce erratic accelerations, we enforce a constant cruise speed ($1.0$\,m/s) to dynamically calculate the corresponding flight duration. The UAV is then driven via the \texttt{moveByVelocityAsync} interface offered within the AirSim environment. This explicit velocity-duration mapping prevents abrupt kinematic transitions and minimizes camera motion blur, ensuring stable, high-quality visual observations for subsequent autoregressive inference steps.

\subsection{Training Objective}
We formulate the learning process as a Behavior Cloning (BC) problem. Given a dataset of expert demonstrations $\mathcal{D} = \{(I_t, P, a^*_t)\}$, where $I_t$ represents the visual observation, $P$ the structured language prompt, and $a^*_t$ the corresponding ground-truth expert action, we optimize the model to minimize the negative log-likelihood of the expert action tokens. The autoregressive training objective is defined as:
\begin{equation}
    \mathcal{L} = - \mathbb{E}_{(I_t, P, a^*_t) \sim \mathcal{D}} \left[ \sum_{k \in \{x, z, \psi\}} \log p(a^*_{t,k} \mid I_t, P, a^*_{t,<k}) \right]
\label{eq:loss}
\end{equation}
where $a^*_{t,k}$ denotes the discrete token for the $k$-th dimension of the expert action. This straightforward frame-level supervision perfectly aligns with our reactive navigation policy.

\section{Experiments}
\label{sec:experiments}

\subsection{Dataset and Evaluation Metrics}
\label{sec:dataset}

\textbf{Dataset.} We evaluate AerialVLA on the TravelUAV benchmark~\cite{wang2025towards}, specifically the \emph{UAV-Need-Help} task, which contains $\sim$12k human-piloted trajectories. Departing from the original oracle-assisted setting, we evaluate purely autonomous navigation guided exclusively by fuzzy directional prompts. We strictly adhere to the official data splits, training on 7,922 trajectories and evaluating across three distinct test sets comprising 1,418 trajectories for the Seen split, 629 for Unseen Object, and 958 for Unseen Map.

\textbf{Evaluation Metrics.} Following standard protocols~\cite{wang2025towards}, we report Navigation Error (NE) measuring the Euclidean distance to the target, Success Rate (SR) indicating the percentage of episodes ending within 20 meters of the target, Oracle Success Rate (OSR), and Success weighted by Path Length (SPL) to balance success with trajectory efficiency.

\subsection{Implementation Details}
\label{sec:implementation}

\textbf{Model and Training Setup.} We instantiate AerialVLA using the OpenVLA-7B backbone~\cite{kim2025openvla}, training on 420k frames from 7,922 expert trajectories. We apply LoRA~\cite{hu2022lora} ($r=64$, $\alpha=128$, dropout $0.05$) to the language backbone, yielding $\sim$2.98\% trainable parameters. Both visual encoders remain frozen, while the projector is fully fine-tuned. Optimization uses AdamW~\cite{loshchilov2019decoupled} (weight decay $0.03$, gradient clip $1.0$) with a cosine scheduler (peak LR $2 \times 10^{-4}$, 5\% warmup). Training runs on 4$\times$ RTX 4090 (24GB) GPUs with a global batch size of 64 for 5 epochs ($\sim$35 hours) in BF16 precision.

\textbf{Simulation Configuration.} 
To strictly align with the decoupled action space defined in Sec.~\ref{sec:action}, we configure the AirSim~\cite{Shah2018AirSim} simulator to operate in \texttt{MaxDegreeOfFreedom} mode rather than the standard \texttt{ForwardOnly} mode. This enables the agent to execute independent yaw rotation ($\Delta \psi$) alongside a forward velocity vector, providing the holonomic agility essential for reactive obstacle avoidance.

\textbf{Computational Efficiency.} Evaluated on an RTX 4090, AerialVLA requires 17GB VRAM and 0.38s total latency, outperforming the 20GB and 0.63s of TravelUAV. While our dual-vision VLA model takes 0.35s versus their 0.26s backbone and prediction head, completely eliminating their 0.37s Assist and Grounding DINO modules drastically reduces system latency. With fuzzy directional hints adding merely 0.03s, our framework offers a vastly faster and more compact architecture for real-time navigation.

\subsection{Baselines}
To evaluate our pure end-to-end paradigm, we compare AerialVLA against three baseline categories. Unlike our single autoregressive stream, these methods inherently rely on specialized decoder heads or external auxiliary modules. All baseline results are directly sourced from their original publications~\cite{wang2025towards, jiang2025longfly} using identical data splits.

\textbf{Heuristic Baselines.} Random Action and Fixed Action serve as lower bounds to quantify task difficulty.

\textbf{Hybrid VLM Approaches.} We evaluate CMA~\cite{Anderson_2018_CVPR}, a traditional recurrent neural network baseline. Notably, we include TravelUAV~\cite{wang2025towards}, which combines a language model feature extractor with hierarchical decoders for trajectory regression, while relying on an external Grounding DINO~\cite{liu2024grounding} for target detection.

\textbf{Recent Generalist Models.} NavFoM~\cite{zhang2026embodied} is a foundation model trained across various embodiments using a specialized trajectory planning head, and LongFly~\cite{jiang2025longfly} leverages explicit spatiotemporal encoding modules.

\definecolor{lightpurple}{RGB}{235, 235, 250} 

\begin{table}[htbp]
\centering
\caption{Comparison on the Test Seen Set. SR, OSR, and SPL are reported in percentage (\%). Bold and underline denote the best and second-best model results.}
\label{tab:comparison_seen}
\setlength{\tabcolsep}{3pt}
\resizebox{\textwidth}{!}{%
\begin{tabular}{lcccccccccccc}
\toprule
\multirow{2}{*}{Method} & \multicolumn{4}{c}{\textbf{Full}} & \multicolumn{4}{c}{\textbf{Easy}} & \multicolumn{4}{c}{\textbf{Hard}} \\
\cmidrule(lr){2-5} \cmidrule(lr){6-9} \cmidrule(lr){10-13}
 & NE$\downarrow$ & SR$\uparrow$ & OSR$\uparrow$ & SPL$\uparrow$ & NE$\downarrow$ & SR$\uparrow$ & OSR$\uparrow$ & SPL$\uparrow$ & NE$\downarrow$ & SR$\uparrow$ & OSR$\uparrow$ & SPL$\uparrow$ \\
\midrule
\rowcolor{gray!10} 
Human & 14.15 & 94.51 & 94.51 & 77.84 & 11.68 & 95.44 & 95.44 & 76.19 & 17.16 & 93.37 & 93.37 & 79.85 \\
\midrule
Random Action     & 222.20 & 0.14  & 0.21  & 0.07  & 142.07 & 0.26  & 0.39  & 0.13  & 320.12 & 0.00  & 0.00  & 0.00  \\
Fixed Action      & 188.61 & 2.27  & 8.16  & 1.40  & 121.36 & 3.48  & 11.48 & 2.14  & 270.69 & 0.79  & 4.09  & 0.49  \\
CMA \cite{Anderson_2018_CVPR}           & 135.73 & 8.37  & 18.72 & 7.90  & 84.89  & 11.48 & 24.52 & 10.68 & 197.77 & 4.57  & 11.65 & 4.51  \\
TravelUAV-DA \cite{wang2025towards}    & 98.66  & 17.45 & 48.87 & 15.76 & 66.40  & 20.26 & 51.23 & 18.10 & 138.04 & 14.02 & 45.98 & 12.90 \\
NavFoM \cite{zhang2026embodied}       & 93.05  & 29.17 & 49.24 & 25.03 & 58.98  & 32.91 & 53.16 & 27.87 & 143.83 & 23.58 & 43.40 & 20.80 \\
LongFly \cite{jiang2025longfly} & \textbf{60.02} & \underline{36.39} & \textbf{65.87} & \underline{31.07} & \textbf{38.10} & \underline{38.52} & \textbf{71.90} & \underline{31.24} & \textbf{85.20} & \underline{33.94} & \textbf{58.94} & \underline{30.88} \\
\midrule
\rowcolor{lightpurple}
\textbf{Ours} & \underline{65.88} & \textbf{47.96} & \underline{57.69} & \textbf{38.54} & \underline{43.76} & \textbf{49.30} & \underline{61.30} & \textbf{37.14} & \underline{93.16} & \textbf{46.30} & \underline{53.23} & \textbf{40.26} \\
\bottomrule
\multicolumn{13}{l}{}
\end{tabular}%
}
\end{table}

\subsection{Quantitative Results}
Tables~\ref{tab:comparison_seen}, \ref{tab:comparison_unseen_obj}, and \ref{tab:comparison_unseen_map} summarize our evaluation across the Seen, Unseen Object, and Unseen Map splits, demonstrating the distinct advantages of our fully autonomous paradigm over assistant-reliant baselines.

\textbf{Performance on Seen Environments.}
As shown in Table~\ref{tab:comparison_seen}, AerialVLA achieves a new state-of-the-art \textbf{47.96\% SR} and \textbf{38.54\% SPL}, surpassing the strongest baseline (LongFly) by substantial margins (+11.57\% SR and +7.47\% SPL). The performance gap becomes even more pronounced on the \textit{Hard} split involving long-horizon flights, where our SR advantage widens to +12.36\%. While LongFly achieves a higher OSR by incorporating ground-truth directional hints, its precipitous drop from OSR (65.87\%) to SR (36.39\%) exposes a critical failure in the final termination phase. In contrast, AerialVLA demonstrates superior OSR-to-SR conversion efficiency. By unifying navigation and landing with an intrinsic stopping mechanism, our agent autonomously executes precise terminal maneuvers without relying on external oracle triggers.

\definecolor{lightpurple}{RGB}{235, 235, 250} 

\begin{table}[htbp]
\centering
\caption{Comparison on the Test Unseen Object Set. SR, OSR, and SPL are reported in percentage (\%). Bold and underline denote the best and second-best results, respectively.}
\label{tab:comparison_unseen_obj}
\setlength{\tabcolsep}{3pt}
\resizebox{\textwidth}{!}{%
\begin{tabular}{lcccccccccccc}
\toprule
\multirow{2}{*}{Method} & \multicolumn{4}{c}{\textbf{Full}} & \multicolumn{4}{c}{\textbf{Easy}} & \multicolumn{4}{c}{\textbf{Hard}} \\
\cmidrule(lr){2-5} \cmidrule(lr){6-9} \cmidrule(lr){10-13}
 & NE$\downarrow$ & SR$\uparrow$ & OSR$\uparrow$ & SPL$\uparrow$ & NE$\downarrow$ & SR$\uparrow$ & OSR$\uparrow$ & SPL$\uparrow$ & NE$\downarrow$ & SR$\uparrow$ & OSR$\uparrow$ & SPL$\uparrow$ \\
\midrule
Random Action     & 260.14 & 0.16  & 0.16  & 0.16  & 174.10 & 0.48  & 0.48  & 0.48  & 302.96 & 0.00  & 0.00  & 0.00  \\
Fixed Action      & 212.84 & 3.66  & 9.54  & 2.16  & 151.66 & 6.70  & 13.88 & 3.72  & 243.29 & 2.14  & 7.38  & 1.38  \\
CMA \cite{Anderson_2018_CVPR}           & 155.79 & 9.06  & 16.06 & 8.68  & 102.92 & 14.83 & 22.49 & 13.90 & 182.09 & 6.19  & 12.86 & 6.08  \\
TravelUAV \cite{wang2025towards}    & 118.11 & 22.42 & 46.90 & 20.51 & 86.12  & 24.40 & 49.28 & 22.03 & 134.03 & 21.43 & 45.71 & 19.75 \\
NavFoM \cite{zhang2026embodied}       & 108.04 & 29.83 & 47.99 & 27.20 & 70.51  & 32.54 & 50.72 & 29.54 & 133.01 & 28.03 & 46.18 & 25.64 \\
LongFly \cite{jiang2025longfly} & \underline{66.74} & \underline{43.87} & \underline{64.56} & \underline{38.39} & \underline{54.84} & \underline{38.01} & \underline{56.84} & \underline{31.36} & \textbf{57.07} & \underline{50.25} & \textbf{74.16} & \underline{45.27} \\
\midrule
\rowcolor{lightpurple}
\textbf{Ours} & \textbf{61.45} & \textbf{56.60} & \textbf{64.86} & \textbf{46.61} & \textbf{45.72} & \textbf{56.94} & \textbf{64.11} & \textbf{43.76} & \underline{69.27} & \textbf{56.43} & \underline{65.24} & \textbf{48.03} \\
\bottomrule
\end{tabular}%
}
\end{table}

\textbf{Generalization to Unseen Objects.}
Table~\ref{tab:comparison_unseen_obj} evaluates robustness against novel target categories, where AerialVLA maintains superiority with \textbf{56.60\% SR} overall. While baselines achieve high OSR on the \textit{Hard} split via oracle guidance, their reliance on explicit object detectors severely limits their ability to recognize and stop at out-of-distribution targets. In contrast, our method directly grounds novel visual concepts to control actions. This confirms that our minimalist framework leverages the open-vocabulary representations inherent in the LLM to identify unseen targets, rather than merely overfitting to training categories.

\textbf{Adaptability to Unseen Maps.}
The Unseen Map Test Set (Table~\ref{tab:comparison_unseen_map}) provides the strongest evidence of our generalization capability. In entirely novel environments, LongFly suffers a drastic degradation to 11.27\% SR due to its heavy reliance on historical context and map-specific priors. In contrast, AerialVLA demonstrates remarkable zero-shot adaptability, achieving \textbf{37.58\% SR} and \textbf{28.22\% SPL}. Both metrics are approximately three times those of the SOTA baseline. This indicates that AerialVLA acquires a fundamental visual servoing capability that seamlessly transfers to novel geometries. By strictly relying on instantaneous observations rather than accumulated spatial memory, our reactive approach exhibits superior robustness against severe environmental shifts.

\definecolor{lightpurple}{RGB}{235, 235, 250} 

\begin{table}[htbp]
\centering
\caption{Comparison on the Test Unseen Map Set. SR, OSR, and SPL are reported in percentage (\%). Bold and underline denote the best and second-best results, respectively.}
\label{tab:comparison_unseen_map}

\setlength{\tabcolsep}{3pt} 

\resizebox{\textwidth}{!}{%
\begin{tabular}{lcccccccccccc}
\toprule
\multirow{2}{*}{Method} & \multicolumn{4}{c}{\textbf{Full}} & \multicolumn{4}{c}{\textbf{Easy}} & \multicolumn{4}{c}{\textbf{Hard}} \\
\cmidrule(lr){2-5} \cmidrule(lr){6-9} \cmidrule(lr){10-13}
 & NE$\downarrow$ & SR$\uparrow$ & OSR$\uparrow$ & SPL$\uparrow$ & NE$\downarrow$ & SR$\uparrow$ & OSR$\uparrow$ & SPL$\uparrow$ & NE$\downarrow$ & SR$\uparrow$ & OSR$\uparrow$ & SPL$\uparrow$ \\
\midrule
Random Action     & 202.98 & 0.00  & 0.00  & 0.00  & 158.46 & 0.00  & 0.00  & 0.00  & 265.88 & 0.00  & 0.00  & 0.00  \\
Fixed Action      & 180.47 & 0.52  & 2.61  & 0.39  & 132.89 & 0.89  & 4.28  & 0.67  & 247.72 & 0.00  & 0.25  & 0.00  \\
CMA \cite{Anderson_2018_CVPR}           & 141.68 & 2.30  & 10.02 & 2.16  & 102.29 & 3.57  & 14.26 & 3.33  & 197.35 & 0.50  & 4.03  & 0.50  \\
TravelUAV \cite{wang2025towards}    & 138.80 & 4.18  & 20.77 & 3.84  & 102.94 & 4.63  & 22.82 & 4.24  & 189.46 & 3.53  & 17.88 & 3.28  \\
NavFoM \cite{zhang2026embodied}       & 125.10 & 6.30  & 18.95 & 5.68  & 102.41 & 6.77  & 20.07 & 6.04  & 170.58 & 5.36  & 15.71 & 4.97  \\
LongFly \cite{jiang2025longfly} & \underline{108.32} & \underline{11.27} & \underline{30.27} & \underline{9.32} & \underline{78.56} & \underline{12.96} & \underline{34.31} & \underline{10.32} & \underline{148.10} & \underline{9.02} & \underline{24.88} & \underline{7.98} \\
\midrule
\rowcolor{lightpurple}
\textbf{Ours} & \textbf{67.42} & \textbf{37.58} & \textbf{52.92} & \textbf{28.22} & \textbf{44.99} & \textbf{41.89} & \textbf{58.47} & \textbf{29.72} & \textbf{99.11} & \textbf{31.49} & \textbf{45.09} & \textbf{26.11} \\
\bottomrule
\end{tabular}%
}
\end{table}

\subsection{Qualitative Analysis}
\begin{figure}[tb]
  \centering
  \includegraphics[width=\linewidth]{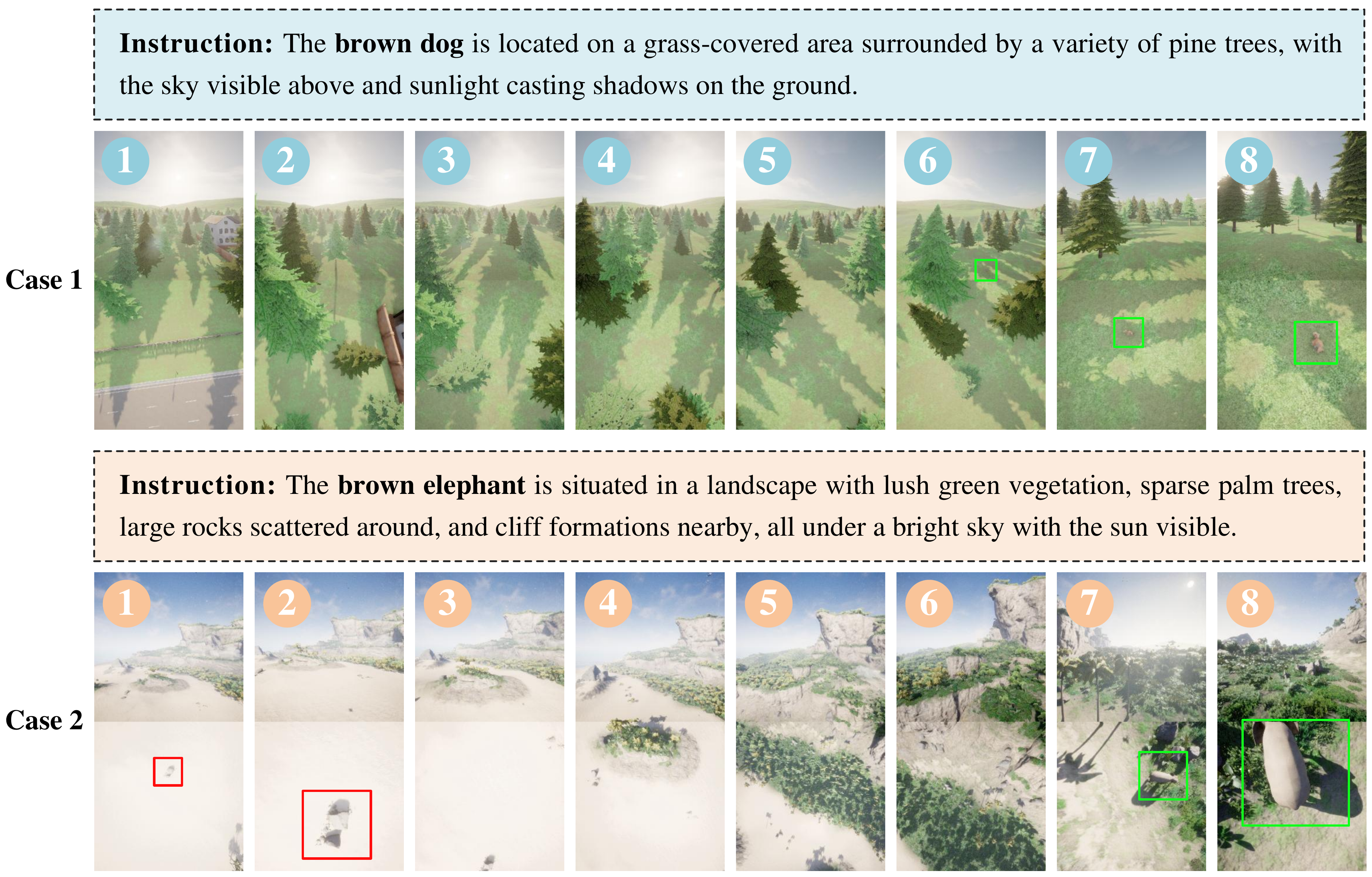}
  \caption{\textbf{Qualitative visualization of our proposed AerialVLA.} We display the vertical mosaic inputs (Front/Down) at key timesteps. The agent demonstrates \textit{precision maneuvering in clutter} (Top) and \textit{active error correction against distractors} (Bottom), validating the robustness of the end-to-end policy.}
  \label{fig:figure_qualitative}
\end{figure}

Figure~\ref{fig:figure_qualitative} visualizes representative trajectories highlighting two distinct navigation behaviors, demonstrating robust decision-making capabilities of the agent.

\textbf{Precision Maneuvering in Clutter.} In unstructured settings like extensive forests (Figure~\ref{fig:figure_qualitative}, Top), AerialVLA leverages its decoupled 3-DoF action space for fine-grained control. The agent horizontally aligns with the target while maintaining altitude above vegetation, followed by a precise vertical descent to safely land.

\textbf{Active Error Correction against Distractors.} Under visual ambiguity (Figure~\ref{fig:figure_qualitative}, Bottom), the agent approaches a distractor object, briefly hovers for inspection, and upon recognizing the mismatch, autonomously climbs to resume the search. This active perception loop proves a capacity for self-correction that strictly exceeds simple trajectory regression.

\subsection{Ablation Study}
We conduct ablation experiments to validate the individual contributions of our core architectural designs, with quantitative results summarized in Table~\ref{tab:ablation_data}.

\textbf{Robustness to Raw Demonstrations.} To verify that our performance gains stem from the architecture rather than dataset curation, we trained a variant on raw, uncleaned demonstrations. While noisy label conflicts predictably cause a slight degradation (e.g., 5.43\% SR drop on unseen maps), this raw variant still achieves 32.15\% SR, preserving a nearly threefold advantage over LongFly. This confirms the inherent robustness of the proposed architecture, proving that geometry-consistent filtering primarily serves to resolve mathematical ambiguities rather than artificially boosting baseline performance.

\textbf{Efficacy of Minimalist Perception.} A five-view variant incorporating redundant side and rear cameras severely degrades performance, plummeting the unseen map SR from 37.58\% to 21.71\%. This substantial decline corroborates our hypothesis that excessive visual inputs distract the agent and induce overfitting to background clutter, empirically validating our streamlined dual-view design.

\textbf{Advantage of Numerical Tokenization.} A variant utilizing custom action tokens instead of standard numerical digits experiences significant SR and SPL drops across all splits. This performance collapse validates our analysis in Sec.~\ref{sec:action}: training novel embeddings from scratch introduces a severe cold-start problem, whereas adopting a pure numerical format leverages the pre-trained magnitude awareness of the LLM to ensure reliable, fine-grained control.

\definecolor{lightpurple}{RGB}{235, 235, 250}

\begin{table}[tbp]
\centering
\caption{Ablation on data curation, perception views, and action tokenization. We compare raw noisy data, 5-view input, and custom action token variants against our default formulation. Metrics are in percentage (\%).}
\label{tab:ablation_data}
\setlength{\tabcolsep}{6pt} 
\resizebox{\textwidth}{!}{%
\begin{tabular}{lcccccc}
\toprule
\multirow{2}{*}{Training Data / Method} & \multicolumn{2}{c}{\textbf{Seen}} & \multicolumn{2}{c}{\textbf{Unseen Object}} & \multicolumn{2}{c}{\textbf{Unseen Map}} \\
\cmidrule(lr){2-3} \cmidrule(lr){4-5} \cmidrule(lr){6-7}
 & SR$\uparrow$ & SPL$\uparrow$ & SR$\uparrow$ & SPL$\uparrow$ & SR$\uparrow$ & SPL$\uparrow$ \\
\midrule
Baseline LongFly~\cite{jiang2025longfly} & 36.39 & 31.07 & 43.87 & 38.39 & 11.27 & 9.32 \\
\midrule
Ours - custom tokens & 39.84 & 31.73 & 38.79 & 32.77 & 26.51 & 19.98 \\
Ours - 5-view input & 41.54 & 32.69 & 51.51 & 42.33 & 21.71 & 13.46 \\
Ours w/o filtering & 40.90 & 32.59 & 51.99 & 42.67 & 32.15 & 22.67 \\
\rowcolor{lightpurple}
\textbf{Ours} & \textbf{47.96} & \textbf{38.54} & \textbf{56.60} & \textbf{46.61} & \textbf{37.58} & \textbf{28.22} \\
\bottomrule
\end{tabular}%
}
\end{table}

\section{Limitations and Future Work}
\label{sec:limitations}
While highly effective, the minimalist design of AerialVLA presents specific avenues for future refinement. First, by prioritizing instantaneous reactive control over explicit historical memory, the agent occasionally faces challenges with global backtracking in highly repetitive urban structures (e.g., the \textit{NewYorkCity} map). Second, bounded by the nature of behavior cloning, the policy acts conservatively in extreme out-of-distribution scenarios. For instance, when targets are severely occluded by dense canopies in the unseen \textit{ModularPark} map, the agent defaults to safe hovering rather than executing complex multi-stage exploration. To address these trade-offs, future work will integrate lightweight memory mechanisms for global reasoning and explore reinforcement learning fine-tuning to enable active exploration beyond static expert demonstrations.

\section{Conclusion}
This paper presents AerialVLA to fundamentally rethink the prevailing modular methodologies in UAV vision-language navigation. Our work introduces a novel perspective to the field by establishing a minimalist end-to-end Vision-Language-Action paradigm. By navigating solely via onboard fuzzy hints and unifying cruising with precise termination into a single autonomous policy, we completely decouple the agent from dense oracle guidance and explicit object detectors. Extensive evaluations prove that discarding redundant spatial memory and complex auxiliary modules inherently fosters exceptional zero-shot generalization across unseen targets and novel map geometries. We anticipate that this pure end-to-end philosophy will serve as a robust foundation for future natively intelligent aerial agents operating in unconstrained open-world environments.

%
%
\bibliographystyle{splncs04}
\bibliography{main}
\end{document}